\documentclass[10pt,twocolumn,letterpaper]{article}

\usepackage{cvpr}              % To produce the CAMERA-READY version
\usepackage{graphicx}
\usepackage{amsmath}
\usepackage{amssymb}
\usepackage{booktabs}
\usepackage{xcolor}
\usepackage{tikz}
\usetikzlibrary{mindmap} %for mindmap

\usepackage[accsupp]{axessibility} % Improves PDF readability for those with disabilities.

\definecolor{JINZAMOMI}{RGB}{225,122,119} %red1
\definecolor{AKEBONO}{RGB}{241,148,131} %red2
\definecolor{TOKI}{RGB}{238,169,169} %pink2
\definecolor{MOGEI}{RGB}{123,162,63} %green1
\definecolor{HIWA}{RGB}{190,194,63} %green2
\definecolor{NOSHIMEHANA}{RGB}{43,95,117}
\definecolor{KUCHINASHI}{RGB}{246,197,85} %Yellow1
\definecolor{USUKI}{RGB}{250,214,137} %Yellow2
\definecolor{HANAASAGI}{RGB}{30,136,168} %blue
\definecolor{SORA}{RGB}{88,178,220} %blue textcolor
\definecolor{NAE}{RGB}{134,193,102} %green textcolor
\definecolor{KOKE}{RGB}{131,138,45} %green!!!
\definecolor{WASURENAGUSA}{RGB}{125,185,222} %blue2
\definecolor{GUNJYO}{RGB}{81,168,221} %blue3
\definecolor{KAMENOZOKI}{RGB}{165,222,228} %blue4

\usepackage[pagebackref,breaklinks,colorlinks]{hyperref}

\usepackage[capitalize]{cleveref}
\crefname{section}{Sec.}{Secs.}
\Crefname{section}{Section}{Sections}
\Crefname{table}{Table}{Tables}
\crefname{table}{Tab.}{Tabs.}

\usepackage{multirow}
\usepackage{fontawesome}
\usepackage{colortbl}

\usepackage{tcolorbox}
\usepackage{xpatch}
\usepackage{diagbox}
\usepackage{utfsym}

\usepackage{amsmath}
\usepackage{amsfonts}
\usepackage{amssymb}
\usepackage{mathrsfs}

\usepackage[T1]{fontenc} 
\pdfmapline{+delphine < Delphine.ttf <T1-WGL4.enc}
\newcommand\delphinefont[1]{{\usefont{T1}{delphine}{m}{n} #1 }}

\usepackage{MnSymbol}
\usepackage{makecell}
\def\cvprPaperID{49} % *** Enter the CVPR Paper ID here
\def\confName{CVPR}
\def\confYear{2023}

\begin{document}

%\title{Text-to-X and X-to-Image Applications: A Survey} %A Survey for AIGC
\title{Vision + Language Applications: A Survey} %Vision + Language Applications Base and Beyond: A Survey  %Fundamentals and Recent Developments

\author{Yutong Zhou, Nobutaka Shimada\\
Ritsumeikan University, Shiga, Japan\\
\href{mailto:zhou@i.ci.ritsumei.ac.jp}{\textcolor{black}{\tt\small zhou@i.ci.ritsumei.ac.jp}}
}

\maketitle

%%%%%%%%% ABSTRACT
\begin{abstract}
Text-to-image generation has attracted significant interest from researchers and practitioners in recent years due to its widespread and diverse applications across various industries. Despite the progress made in the domain of vision and language research, the existing literature remains relatively limited, particularly with regard to advancements and applications in this field. This paper explores a relevant research track within multimodal applications, including text, vision, audio, and others. In addition to the studies discussed in this paper, we are also committed to continually updating the latest relevant papers, datasets, application projects and corresponding information at \href{https://github.com/Yutong-Zhou-cv/Awesome-Text-to-Image}{https://github.com/Yutong-Zhou-cv/Awesome-Text-to-Image}.
%\url{https://github.com/Yutong-Zhou-cv/Awesome-Text-to-Image}. 
%For double-blind
%In addition to the studies discussed in this paper, we also commit to continually updating relevant papers, datasets, and projects. Corresponding information will be made publicly available on GitHub.

%In recent years, text-to-image generation has garnered highly considerable interest and attention from researchers and practitioners due to its widespread and diverse applications across various industries. However, the existing literature on this research field, particularly concerning the advancements and applications of vision and language, is still relatively scarce, necessitating more comprehensive and in-depth survey literature. This paper presents a deep dive into a pertinent research track within multimodal applications ranging from text, vision, audio and others. In addition to the studies discussed in this article, we are also committed to continually updating the latest relevant papers, datasets, application projects and corresponding information at \url{https://github.com/Yutong-Zhou-cv/Awesome-Text-to-Image}.
\end{abstract}

%%%%%%%%% BODY TEXT
%#########################################################################
\section{Introduction}
\label{sec:intro}

\begin{quotation}
\textit{“The baby, assailed by eyes, ears, nose, skin, and entrails at once, feels it all as one great blooming, buzzing confusion.”}

\small \rightline{-- William James (1890)}
\end{quotation}

The human perceptual system is a complex and multifaceted construct. The five basic senses of hearing, touch, taste, smell, and vision serve as primary channels of perception, allowing us to perceive and interpret most of the external stimuli encountered in this “blooming, buzzing confusion” world. These stimuli always come from multiple events spread out spatially and temporally distributed. The human brain processes conceptual representation based on various modes of perception, with visual information accounting for as much as 87$\%$. Furthermore, in cases where certain aspects of content cannot be only conveyed through visual information, combining multiple sensory modalities can potentially provide a more comprehensive understanding of concepts.
%There are five basic human senses: touch, taste, smell, hearing and vision. Each sensory modality helps us perceive and understand the “blooming, buzzing confusion” world, which continuously challenges our perceptual system with various stimuli such as light, sound, heat, fragrance, pressure, \etc. The human brain processes conceptual representation based on various modes of perception, with visual information accounting for as much as 87$\%$. Furthermore, in cases where certain aspects of content cannot be only conveyed through visual information, combining and converting vision, language and other information can provide complementary information, potentially supplying a more comprehensive expression of concepts. 
In other words, we constantly perceive the world in a “multimodal” manner, which combines different information channels to distinguish features within confusion, seamlessly integrates various sensations from multiple modalities and obtains knowledge through our experiences.

\begin{figure}[!t]
\centering
%{\fontfamily{qag}\selectfont
\tikzstyle{level 2 concept}+=[sibling angle=-47]
\tikzstyle{level 3 concept}+=[sibling angle=33]
\resizebox{1\columnwidth}{!}{ %
\begin{tikzpicture}[scale=1, transform shape]
\path[mindmap,concept color=black,text=white]
node[concept] {\Large \delphinefont{\textbf{Vision + Language \\Multimodal Learning}} \\ \vspace{2mm}\includegraphics[width=1.9cm,height=1.9cm]{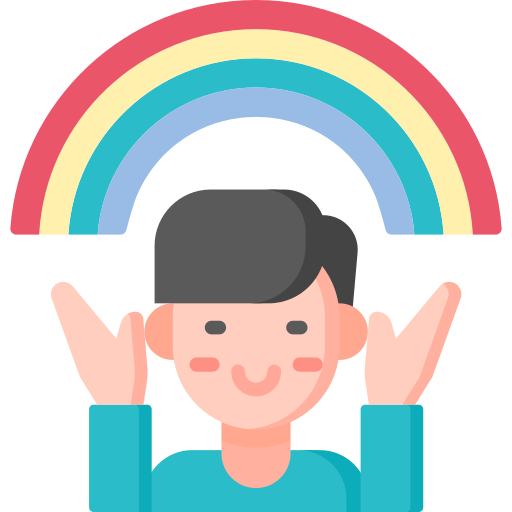}} %[clockwise from=35]
child[concept color=KUCHINASHI,grow=-170]{ 
node[concept] {\delphinefont{\Large Background} (\cref{sec:background})} [clockwise from=155]
child {node[concept] {Dataset (\cref{sec:dataset})}}
%child {node[concept] {Evaluation Metrics (\cref{sec:eval})}}
child {
node[concept] {Evaluation Metrics (\cref{sec:eval})}[clockwise from=-103]
child {node [concept,color=USUKI,text=black]{{\textit{\scriptsize Human \\ Evaluation \\ \hyperref[HE]{\large $\filledstar$}}}}}
child {node [concept,color=USUKI,text=black]{{\textit{\scriptsize Automatic \\ Evaluation \\ \hyperref[AE]{\large $\filledstar$}}}}}
}
}
child[concept color=HANAASAGI, grow=-133]{ 
node[concept] {\delphinefont{\Large Generative Models} (\cref{sec:model})} [clockwise from=188]
child {node[concept] {GAN-based Model (\cref{sec:gan-model})}}
child {node[concept] {Diffusion-based Model (\cref{sec:diff-model})}}}
child[concept color=MOGEI, grow=-78]{ 
node[concept] {\delphinefont{\Large Generative Applications} (\cref{sec:application})} [clockwise from=-158]
%child {node[concept] {Text-to-Image (\cref{sec:t2i})}} 
child {
node[concept] {Text-to-Image (\cref{sec:t2i})} [clockwise from=-148]
child {node [concept,color=HIWA,text=black]{{\textit{\scriptsize Text-to-\\Face \\ \hyperref[T2F]{\large $\filledstar$}}}}} %4
} 
%child {node[concept] {Text-to-X (\cref{sec:t2x})}}
child {
node[concept] {Text-to-X (\cref{sec:t2x})} [clockwise from=-100]
%child {node [concept,color=HIWA,text=white]{Music}} %5
child {node [concept,color=HIWA,text=black]{{\textit{\scriptsize Human Motion \\ \hyperref[motion]{\large $\filledstar$}}}}} %4
% child {node [concept,color=HIWA,text=black]{{\textit{\scriptsize Shape \\ \hyperref[shape]{\large $\filledstar$}}}}} %3
child {node [concept,color=HIWA,text=black]{{\textit{\scriptsize 3D \\ \hyperref[3d]{\large $\filledstar$}}}}} %2
child {node [concept,color=HIWA,text=black]{{\textit{\scriptsize Video \\ \hyperref[video]{\large $\filledstar$}}}}} %1
}
%child {node[concept] {X-to-Image (\cref{sec:x2i})}}
child {
node[concept] {X-to-Image (\cref{sec:x2i})} [clockwise from=-17]
child {node [concept,color=HIWA,text=black]{{\textit{\scriptsize Speech /\\Sound \\ \hyperref[speech]{\large $\filledstar$}}}}} %4
child {node [concept,color=HIWA,text=black]{{\textit{\scriptsize Brain \\ \hyperref[brain]{\large $\filledstar$}}}}} %3
child {node [concept,color=HIWA,text=black]{{\textit{\scriptsize Layout \\ \hyperref[layout]{\large $\filledstar$}}}}} %2
child {node [concept,color=HIWA,text=black]{{\textit{\scriptsize Text + Image \\ \hyperref[t_plus_i]{\large $\filledstar$}}}}} %1
%child {node [concept,color=HIWA,text=white]{Text + Audio + Image}}
}
%child {node[concept] {Downstream Tasks (\cref{sec:downstream})}}
%%%% Optional
% child {
% node[concept] {Downstream Tasks (\cref{sec:downstream})} [clockwise from=33]
% child {node [concept,color=HIWA,text=black]{\textit{{\Large ...}}}}
% child {node [concept,color=HIWA,text=black]{{\textit{\scriptsize Style \\Transfer \\ \hyperref[ST]{\large $\filledstar$}}}}}
% child {node [concept,color=HIWA,text=black]{{\textit{\scriptsize Visual \\Retrieval \\ \hyperref[VR]{\large $\filledstar$}}}}}
% }
child {node[concept] {Multi Tasks (\cref{sec:multi})}}
% child {
% node[concept] {Multi Tasks (\cref{sec:multi})} [clockwise from=-31]
% child {node [concept,color=HIWA,text=black]{{\textit{ Downstream \\Tasks\\ \hyperref[downstream]{\large $\filledstar$}}}}}
% }
}
%
% child[concept color=JINZAMOMI, grow=-8]{  
% node[concept] {\delphinefont{\Large Discussion} (\cref{sec:discussion})} [clockwise from=-95]
% child {node[concept] {Ethical Considerations (\cref{sec:ethical})}}
% %child {node[concept] {Compounding Issues (\cref{sec:issue})}}
% child {
% node[concept] {Compounding Issues (\cref{sec:issue})} [clockwise from=-50]
% child {node [concept,color=TOKI,text=white]{\textbf{\textit{Business Analysis \\ \hyperref[BA]{\large $\filledstar$}}}}}
% child {node [concept,color=TOKI,text=white]{\textbf{\textit{Prompt Engineering \\ \hyperref[PE]{\large $\filledstar$}}}}}
% child {node [concept,color=TOKI,text=white]{\textbf{\textit{Computational Aesthetic \\ \hyperref[AQ]{\large $\filledstar$}}}}}
% }
% child {node[concept] {Limitations (\cref{sec:limitations})}}
% child {node[concept] {Future Directions (\cref{sec:future})}}
% };
child[concept color=JINZAMOMI, grow=-10]{  
node[concept] {\delphinefont{\Large Discussion} (\cref{sec:discussion})} [clockwise from=-80]
%child {node[concept] {Ethical Considerations (\cref{sec:ethical})}}
%child {node[concept] {Compounding Issues (\cref{sec:issue})}}
child {
node[concept] {Compounding Issues (\cref{sec:issue})} [clockwise from=-55]
child {node [concept,color=TOKI,text=black]{{\textit{Prompt \\ Engineering \\ \hyperref[PE]{\large $\filledstar$}}}}}
child {node [concept,color=TOKI,text=black]{{\textit{Computational \\ Aesthetic \\ \hyperref[AQ]{\large $\filledstar$}}}}}
}
child {
node[concept] {Business Analysis (\cref{sec:business})} [clockwise from=-46]
child {node [concept,color=TOKI,text=black]{{\textit{Ethical \\ Considerations \\ \hyperref[EC]{\large $\filledstar$}}}}}
child {node [concept,color=TOKI,text=black]{{\textit{\scriptsize Online \\ Platforms \\ \hyperref[BA]{\large $\filledstar$}}}}}
}
% child {node[concept] {Challenges (\cref{sec:limitations})}}
% child {node[concept] {Future Directions (\cref{sec:future})}}
child {node[concept] {\scriptsize Challenges \& Future Outlooks\\(\cref{sec:future})}}
};
\end{tikzpicture}
}
%}
\caption{\textbf{Overall structure of our survey.} Zoom in for details.}
\label{fig:overview}
\vspace{-2mm}
\end{figure}

In the last few decades, Computer Vision (CV) and Natural Language Processing (NLP) have made several major technological breakthroughs in deep learning research. In recent years, in addition to the significant advancements made in single-modality models~\cite{brown2020language,zeng2021pangu,wang2023neural}, there has been an upsurge in research activities focused on developing large-scale multimodal approaches~\cite{ramesh2021zero,fei2022towards,OpenAI2023GPT4TR}.
%The underlying mechanism of deep learning algorithms is a neural network trained to optimize a mathematically defined objective through a loss function. This optimization process utilizes a numerical procedure called gradient descent to minimize the loss. Consequently, deep learning models only handle numerical inputs and produce numerical outputs. However, when it comes to multimodal tasks, the data is often unstructured, such as the most common text or images. There are two primary difficulties in addressing multimodal assignments~\cite{akkus2023multimodal}: (1) Exactly represent the input into a numerical format; (2) Effectively combine diverse modalities.
Despite the impressive advancements in text-to-image synthesis and its related tasks, which effectively incorporate multiple modalities, existing literature surveys~\cite{bodnar2018text,agnese2020survey,frolov2021adversarial,zhou2021survey,zhan2021multimodal,Park2021BenchmarkFC,gozalo2023chatgpt,cao2023comprehensive,zhang2023text} still remain lacking to thoroughly review and summarize the progress of the applications and challenges in this field.
As illustrated in \cref{fig:overview}, this review complements previous surveys on the text-to-image synthesis task with a focus on representative algorithms and expanded applications beyond text-to-image. Our survey aims to accomplish several goals: (1) Present a comprehensive reference that covers the current state-of-the-art research progress and practical breakthroughs in multimodal learning centering on vision and language, (2) Provide a valuable resource for novice researchers seeking to explore this field, (3) Investigate compounding issues and business analysis in Artificial Intelligence-Generated Content (AIGC) field, and (4) Inspire future research directions and upcoming advances through emerging trends and challenges.

%(1) Present a comprehensive reference that covers the latest theoretical and practical breakthroughs in multimodal learning with a focus on vision and language, (2) Provide a valuable resource for novice researchers seeking to explore this field, (3) Investigate ethical concerns related to computer vision technology in creative fields, and (4) Inspire future research directions through emerging trends and challenges.

%(1) Provide cutting-edge progress in multimodal applications for novice researchers, (2) Provide a comprehensive and reliable reference resource for current works in this domain, (3) Investigate ethical concerns related to applying computer vision technology in creative fields, and (4) Inspire future research endeavors through emerging trends and challenges.
%Our survey aims to accomplish several goals: (1) Provide a brief learning path for new researchers, (2) Investigate ethical concerns related to applying computer vision technology in creative fields, (3) Serve as a valuable reference for current related works, (4) Inspire future research directions through trends and challenges, and (5) Help researchers to further advance the field. 

%######################################

\begin{table*}[]
\centering
\fontsize{7}{9.3}\selectfont
\caption{\textbf{Chronological timeline of representative text-to-image datasets.} \textit{“Public”} includes a link to each dataset (if available) or paper (if not). \textit{“Annotations”} denotes the number of text descriptions per image. \textit{“Attrs”} denotes the total number of attributes in each dataset.}
\begin{tabular}{ccccccccc}
\toprule[1.5pt]
 \multirow{2.5}{*}{\small \textbf{Year}} &\multirow{2.5}{*}{\small \textbf{Dataset}} & \multirow{2.5}{*}{\small \textbf{Public}} & \multicolumn{5}{c}{\small \textbf{Details}} \\ \cmidrule{4-8}
                              &              &            & Category  & Images (Resolution) & Annotations & Attrs & Other Information              \\ \hline
\cellcolor{KUCHINASHI!10} 2008& Oxford-102 Flowers~\cite{nilsback2008automated} & \href{https://www.robots.ox.ac.uk/~vgg/data/flowers/102/index.html}{\color{green}\faCheckCircle}  & Flower & 8,189 (-) & 10 & - & - \\
\cellcolor{KUCHINASHI!15} 2011& CUB-200-2011~\cite{wah2011caltech} &  \href{http://www.vision.caltech.edu/datasets/cub_200_2011/}{\color{green}\faCheckCircle} & Bird & 11,788 (-) & 10 & - & BBox, Segmentation... \\
\cellcolor{KUCHINASHI!20} 2014& MS-COCO2014~\cite{lin2014microsoft} & \href{https://cocodataset.org/#overview}{\color{green}\faCheckCircle} & Iconic Objects & 120k (-) & 5 & - & BBox, Segmentation...\\ %\hline
\cellcolor{KUCHINASHI!30} 2018& Face2Text~\cite{gatt2018face2text} & \href{https://github.com/mtanti/face2text-dataset/}{\color{green}\faCheckCircle} & Face & 10,177 (-) & 1$\sim$ & - & -\\
\cellcolor{KUCHINASHI!40} 2019& SCU-Text2face~\cite{chen2019ftgan} & \href{https://arxiv.org/abs/1904.05729}{\color{red}\faTimesCircleO} & Face & 1,000 (256$\times$256) & 5 & - & -\\
\cellcolor{KUCHINASHI!50} 2020& Multi-Modal CelebA-HQ~\cite{xia2021tedigan} &  \href{https://github.com/IIGROUP/MM-CelebA-HQ-Dataset}{\color{green}\faCheckCircle} & Face & 30,000 (512$\times$512) & 10 & 38 & Masks, Sketches \\
\cellcolor{KUCHINASHI!65} 2021& FFHQ-Text~\cite{zhou2021} & \href{https://github.com/Yutong-Zhou-cv/FFHQ-Text_Dataset}{\color{green}\faCheckCircle} & Face & 760 (1024$\times$1024) & 9 & 162 & BBox \\ 
\cellcolor{KUCHINASHI!65} 2021& M2C-Fashion~\cite{zhang2021ufc} & \href{https://proceedings.neurips.cc/paper/2021/hash/e46bc064f8e92ac2c404b9871b2a4ef2-Abstract.html}
{\color{red}\faTimesCircleO} & Clothing  & 10,855,753 (256$\times$256) & 1 & - & - \\
\cellcolor{KUCHINASHI!65} 2021& CelebA-Dialog~\cite{jiang2021talk} & \href{http://mmlab.ie.cuhk.edu.hk/projects/CelebA/CelebA_Dialog.html}{\color{green}\faCheckCircle} & Face & 202,599 (178$\times$218) & $\sim$5 & 5 & Identity Label... \\
\cellcolor{KUCHINASHI!65} 2021& Faces a la Carte~\cite{wang2021faces} & \href{https://openaccess.thecvf.com/content/WACV2021/papers/Wang_Faces_a_la_Carte_Text-to-Face_Generation_via_Attribute_Disentanglement_WACV_2021_paper.pdf}{\color{red}\faTimesCircleO} & Face & 202,599 (178$\times$218) & $\sim$10 & 40 & - \\
\cellcolor{KUCHINASHI!65} 2021& LAION-400M~\cite{schuhmann2021laion} & \href{https://laion.ai/blog/laion-400-open-dataset/}{\color{green}\faCheckCircle} & Random Crawled & 400M (-) & 1 & - & KNN Index... \\ 
\cellcolor{KUCHINASHI!80} 2022& Bento800~\cite{zhou2022able} &  \href{https://github.com/Yutong-Zhou-cv/Bento800_Dataset}{\color{green}\faCheckCircle} & Food & 800 (600$\times$600) & 9 & - & BBox, Segmentation, Label... \\ 
\cellcolor{KUCHINASHI!80} 2022& LAION-5B~\cite{schuhmann2022laion} & \href{https://laion.ai/blog/laion-5b/}{\color{green}\faCheckCircle} & Random Crawled & 5.85B (-) & 1 & - & URL, Similarity, Language... \\ 
\cellcolor{KUCHINASHI!80} 2022& DiffusionDB~\cite{wang2022diffusiondb} &  \href{https://poloclub.github.io/diffusiondb/}{\color{green}\faCheckCircle} & Synthetic Images & 14M (-) & 1 & - & Size, Random Seed... \\ 
\cellcolor{KUCHINASHI!80} 2022& COYO-700M~\cite{kakaobrain2022coyo-700m} & \href{https://github.com/kakaobrain/coyo-dataset}{\color{green}\faCheckCircle} & Random Crawled & 747M (-) & 1 & - & URL, Aesthetic Score... \\ 
\cellcolor{KUCHINASHI!80} 2022& DeepFashion-MultiModal~\cite{jiang2022text2human} & \href{https://github.com/yumingj/DeepFashion-MultiModal}{\color{green}\faCheckCircle} & Full Body & 44,096 (750$\times$1101) & 1 & - & Densepose, Keypoints...\\
\cellcolor{KUCHINASHI} 2023& ANNA~\cite{ramakrishnan2023anna} &  \href{https://github.com/aashish2000/ANNA}{\color{green}\faCheckCircle} & News & 29,625 (256$\times$256) & 1 & - & -\\ 
\cellcolor{KUCHINASHI} 2023& DreamBooth~\cite{ruiz2023dreambooth} &  \href{https://github.com/google/dreambooth}{\color{green}\faCheckCircle} & Objects \& Pets & 158 (-) & 25 & - & -\\ 
\bottomrule[1.5pt]
\end{tabular}
\label{tab:t2i-dataset}
\vspace{-2mm}
\end{table*}

%#########################################################################
\section{Background} 
\label{sec:background}

This section introduces commonly used datasets and evaluation metrics of the text-to-image generation task.
%task formulation

%-------------------------------------------------------------------------
\subsection{Datasets}
\label{sec:dataset}

As one might expect, NLP models primarily rely on textual data for training, while CV models train on image-based information. Vision-language pre-trained models employ a combination of text and images, merging the capabilities of both NLP and CV. Training complex models requires accurate annotation of data by expert human annotators. This process incurs considerable costs and resources, particularly when working with large datasets or domain-specific information.
\Cref{tab:t2i-dataset} presents a comprehensive list of notable datasets, ranging from small-scale~\cite{nilsback2008automated,wah2011caltech,gatt2018face2text,chen2019ftgan,zhou2021,zhou2022able,ramakrishnan2023anna,ruiz2023dreambooth} to large-scale~\cite{lin2014microsoft,zhang2021ufc,jiang2021talk,wang2021faces,xia2021tedigan,schuhmann2021laion,schuhmann2022laion,jiang2022text2human,wang2022diffusiondb,kakaobrain2022coyo-700m}, within the field of text-to-image generation research.

~\cite{nilsback2008automated,wah2011caltech,lin2014microsoft} are widely known in the early stages of text-to-image generation research and regarded as the most commonly utilized dataset in the field.~\cite{zhou2021} is a small-scale collection of face images with extensive facial attributes specifically designed for text-to-face generation and text-guided facial image manipulation.~\cite{zhou2022able} is the first manually annotated synthetic box lunch dataset containing diverse annotations to facilitate innovative aesthetic box lunch presentation design.~\cite{jiang2022text2human} is a high-quality human dataset annotated with rich multi-modal labels, including human parsing labels, keypoints, fine-grained attributes and textual descriptions. As the pioneer in large-scale text-to-image datasets,~\cite{wang2022diffusiondb} comprises 14 million images generated through Stable Diffusion~\cite{rombach2022high} using prompts and hyperparameters provided by real users. Furthermore, ~\cite{schuhmann2022laion} contains 5.85 billion CLIP-filtered image-text pairs, the largest publicly available image-text dataset globally. 

%-------------------------------------------------------------------------
\subsection{Evaluation Metrics}
\label{sec:eval}
When evaluating the performance of text-to-image generation models, two assessments are commonly employed: automated evaluation and human evaluation. 
The former measures image realism and the alignment between generated images and corresponding text descriptions. The latter relies on humans to make subjective judgments.

\textbf{{Automatic Evaluation}}\label{AE}
Automatic evaluation is a widely used approach for assessing the quality of generated images and quantitatively measuring the alignment between the generated images and input textual descriptions. 
%There are several objective metrics used for automatic evaluation, such as Inception Score (IS)~\cite{salimans2016improved}, Fréchet Inception Distance (FID)~\cite{heusel2017gans}, and Structural Similarity (SSIM). 

For image quality assessment, Inception Score (IS)~\cite{salimans2016improved} and Fréchet Inception Distance (FID)~\cite{heusel2017gans} are two commonly utilized metrics. 
IS evaluates the quality and diversity of the generated images. \textit{Higher is better.} FID measures the Fréchet distance between the generated images and the ground truth images. \textit{Lower is better.}

For text-image consistency assessment, R-precision (RP)~\cite{xu2018attngan} evaluates the degree of alignment between a generated image and its corresponding text description. \textit{Higher is better.} Semantic object accuracy (SOA)~\cite{hinz2020semantic} utilizes a pre-trained object detector to check whether the generated image contains objects mentioned in the corresponding textual description. User studies have demonstrated that SOA shows a stronger correlation with human perception than IS. \textit{Higher is better.} Positional Alignment (PA)~\cite{dinh2022tise} evaluates the consistency between the spatial placement of the visual elements within the generated image and their corresponding positions in the text description. \textit{Higher is better.}

Although automatic evaluation metrics have demonstrated effectiveness in assessing the quality of generated images in text-to-image synthesis, these metrics are subject to several limitations. One of the most significant issues is that such metrics may attribute high scores to images that appear realistic but fail to represent the intended meaning of the corresponding textual description accurately. Moreover, automatic metrics typically neglect the subjective nature of human perception, which plays a crucial role in evaluating the quality of generated images. Therefore, complementary human evaluation metrics are necessary to provide a comprehensive assessment of the quality of generated images in text-to-image synthesis.

\textbf{{Human Evaluation}}\label{HE}
To confirm the credibility and correlation of automatic evaluations, several works~\cite{park2021benchmark,saharia2022photorealistic,lin2022magic3d,li2022upainting,petsiuk2022human} have conducted user analysis to evaluate the performance of generated images against human judgments. In ~\cite{dinh2022tise,yu2022scaling}, participants are asked to rate generated images based on two criteria: plausibility (including object accuracy, counting, positional alignment, or image-text alignment) and naturalness (whether the image appears natural or realistic). The evaluation protocol is designed in a 5-Point Likert ~\cite{ref1} manner, in which human evaluators rate each prompt on a scale of 1 to 5, with 5 representing the best and 1 representing the worst. A human evaluation in ~\cite{petsiuk2022human} was conducted for three challenging tasks to compare the performance of Stable Diffusion~\cite{rombach2022high} and DALL-E 2~\cite{ramesh2022hierarchical}. Moreover, for rare object combinations that require common sense understanding or aim to avoid bias related to race or gender, human evaluation is even more important. Such concepts are difficult to define, and human expertise is required to provide accurate evaluations.

%#########################################################################
\section{Generative Models}
\label{sec:model}

%-------------------------------------------------------------------------
\subsection{GAN-based Model}
\label{sec:gan-model}

Generative Adversarial Networks (GANs)~\cite{goodfellow2014generative} have been a significant breakthrough in generative modeling, with the unique ability to generate realistic and diverse samples. GANs consist of two components, the generator and the discriminator, which engage in a continuous adversarial process. The generator produces synthetic images from random noise, while the discriminator aims to distinguish between these generated images and real images from the training dataset. Through backpropagation and optimization, the generator continually refines its outputs in response to the feedback from the discriminator and generates more realistic images. GANs have been applied to many applications, including image generation~\cite{bao2017cvae,karras2020analyzing,kang2023scaling}, super-resolution~\cite{bulat2018learn,zhang2022soup}, 3D object generation~\cite{smith2017improved,zhang2022monocular}, \etc. From image inpainting~\cite{demir2018patch,zhang2022gan} to video generation~\cite{tulyakov2018mocogan,gupta2022rv}, from makeup transfer~\cite{jiang2020psgan,yang2022elegant,xu2022tsev} to virtual try-on~\cite{xie2021towards,lewis2021tryongan}, GANs solve various problems and create new possibilities across multiple industries. 

%-------------------------------------------------------------------------
\subsection{Diffusion Model}
\label{sec:diff-model}

Diffusion models (DMs), also commonly known as diffusion probabilistic models~\cite{sohl2015deep}, represent a class of generative models founded on Markov chains and trained through weighted variational inference~\cite{ho2020denoising}. The primary objective of ~\cite{ho2020denoising} is to learn the impact of noise on the available information in the sample or the degree to which the diffusion process reduces the information available. The two-step process in~\cite{ho2020denoising} consists of forward and reverse diffusion. In the forward diffusion process, Gaussian noise is successively introduced, representing the diffusion process until the data becomes total noise. The reverse diffusion process trains a neural network to learn the conditional distribution probabilities, allowing the model to reverse the noise and reconstruct the original data effectively. DMs also address some challenges associated with GANs, such as mode collapse and training instability. 

%#########################################################################
\section{Generative Applications}
\label{sec:application}

In the previous sections, we explore generative models from the perspective of their background and basic mechanisms. This section delves into the applications in multimodal backgrounds, focusing on their utilization in generating and analyzing data. The discussion is organized into subsections, each focusing on a combination of various modalities, particularly in \cref{sec:t2x} and \cref{sec:x2i}, where “X” represents the additional data. We analyze the significance of integrating multiple modalities for enhanced performance in vision and language tasks. Additionally, we also explore the methodologies utilized in developing these multimodal learning frameworks and discuss their respective contributions to advancing the field.

\subsection{Text-to-Image} 
\label{sec:t2i}

Advancements in the text-to-image generation domain can be broadly classified into three primary categories: GAN-based methods, diffusion-based methods and autoregressive methods, as illustrated in \cref{tab:T2I}. GAN-based approaches are primarily based on stackGAN~\cite{zhang2017stackgan} and styleGAN~\cite{karras2019style} architectures to generate high-quality and visually coherent images. 
Diffusion-based methods model the image generation process as a series of diffusion steps~\cite{sohl2015deep}, progressively refining the generated image. 
Inspired by the success of autoregressive Transformers~\cite{vaswani2017attention}, autoregressive methods focus on sequentially predicting individual pixels or regions in an image, generating the output based on a learned probability distribution. 
Each approach has unique advantages and challenges in text-to-image synthesis, and further research aims to address their limitations and enhance their capabilities.

%StackGAN~\cite{zhang2017stackgan} utilizes a two-stage process, where the first stage generates a low-resolution image based on the textual prompt, and the second stage refines it to produce a high-resolution output. 

\begin{table}[]
\centering
\fontsize{5}{7.5}\selectfont
\caption{\textbf{A comprehensive list of text-to-image approaches.} The pioneering works in each development stage are highlighted in blue. Text-to-face generation works are underlined.}
\begin{tabular}{c||l|l}
\toprule[1.5pt]
\multicolumn{2}{c} {\textbf{Models}}  & \multicolumn{1}{c}{\textbf{Years: Methods}}\\
\midrule
\multirow{4}{*}{\makecell*[c]{GAN\\(~\cref{sec:gan-model} )}} & Conditional GAN~\cite{mirza2014conditional}-based & {\makecell*[l]{\textbf{2016-2021:}\setlength{\fboxsep}{1pt}\colorbox{GUNJYO!140}{\cite{reed2016generative}},\cite{reed2016learning},~\cite{cha2017adversarial},~\cite{dong2017i2t2i},~\cite{hong2018inferring},~\cite{park2018mc},~\cite{el2019tell}}}        \\
                           & StackGAN\setlength{\fboxsep}{1pt}\colorbox{GUNJYO!140}{\cite{zhang2017stackgan}}-based &{\makecell*[l]{\textbf{2018:}~\cite{zhang2018stackgan++},~\cite{gorti2018text},~\cite{xu2018attngan},~\cite{sharma2018chatpainter}\\\textbf{2019:}\underline{~\cite{chen2019ftgan}},~\cite{joseph2019c4synth},~\cite{tan2019semantics},~\cite{yin2019semantics},~\cite{qiao2019mirrorgan},~\cite{li2019controllable},~\cite{zhu2019dm}\\\textbf{2020:}~\cite{zhu2020cookgan},~\cite{cheng2020rifegan},~\cite{tan2020kt},~\cite{liang2020cpgan},~\cite{wang2020end},~\cite{hinz2020semantic}\\\textbf{2021:}\underline{~\cite{sun2021multi}},\underline{~\cite{zhou2021}},\underline{~\cite{zhou2021generative}},~\cite{yang2021multi},~\cite{dong2021unsupervised},~\cite{cheng2021rifegan2}\\\textbf{2022:}\underline{~\cite{luo2022dualg}},\underline{~\cite{luo2022cmafgan}},~\cite{tan2022dr},~\cite{wu2022t} }}  \\
                           & StyleGAN~\cite{karras2019style}-based        & {\makecell*[l]{\textbf{2021:}\setlength{\fboxsep}{1pt}\colorbox{GUNJYO!140}{\underline{~\cite{xia2021tedigan}}},\underline{~\cite{wang2021faces}},~\cite{10.1145/3474085.3475226}\\\textbf{2022:}\underline{~\cite{du2022text}},\underline{~\cite{pinkney2022clip2latent}},\underline{~\cite{hou2022textface}},\underline{~\cite{sun2022anyface}},\underline{~\cite{sabae2022stylet2f}},\underline{~\cite{li2022stylet2i}},~\cite{zhou2021lafite} }}\\
                           & Others &  {\makecell*[l]{\textbf{2018:}~\cite{zhang2018photographic} (Hierarchical adversarial network)\\ \textbf{2021:}~\cite{Liu2021FuseDreamTT,Frans2021CLIPDrawET} (BigGAN~\cite{Brock2018LargeSG})\\
                           \textbf{2022:}~\cite{liao2022text} (One-stage framework)}} \\
\midrule
{\makecell*[c]{Diffusion\\(~\cref{sec:diff-model} )}} & Diffusion~\cite{sohl2015deep}-based &  {\makecell*[l]{\textbf{2022:}\setlength{\fboxsep}{1pt}\colorbox{GUNJYO!140}{~\cite{rombach2022high}},~\cite{gu2022vector},~\cite{ramesh2022hierarchical},~\cite{saharia2022photorealistic},~\cite{nichol2021glide},~\cite{liu2022compositional}\\~\cite{hertz2022prompt},~\cite{wu2022creative},~\cite{kapelyukh2022dall},~\cite{li2022swinv2},~\cite{feng2022ernie},~\cite{balaji2022ediffi},~\cite{kumari2022multi} \\
\textbf{2023:}~\cite{li2023gligen},~\cite{feng2022training},~\cite{chefer2023attend},~\cite{zhang2023adding},~\cite{orgad2023editing} }}\\
\midrule
\tiny Autoregressive   &       Transformer~\cite{vaswani2017attention}-based & {\makecell*[l]{\textbf{2021:}\setlength{\fboxsep}{1pt}\colorbox{GUNJYO!140}{\cite{ramesh2021zero}},\cite{ding2021cogview},~\cite{10.1145/3474085.3481540}\\
\textbf{2022:}~\cite{ding2022cogview2},~\cite{yu2022scaling},~\cite{li2022neural},~\cite{cho2022dall},~\cite{wang2022clip},~\cite{wu2022text},~\cite{lee2022autoregressive},~\cite{gafni2022make}  }}     \\
\bottomrule[1.5pt]
\end{tabular}
\label{tab:T2I}
\vspace{-2mm}
\end{table}

\begin{figure*}[t]
  \centering
  \includegraphics[width=0.95\textwidth,clip]{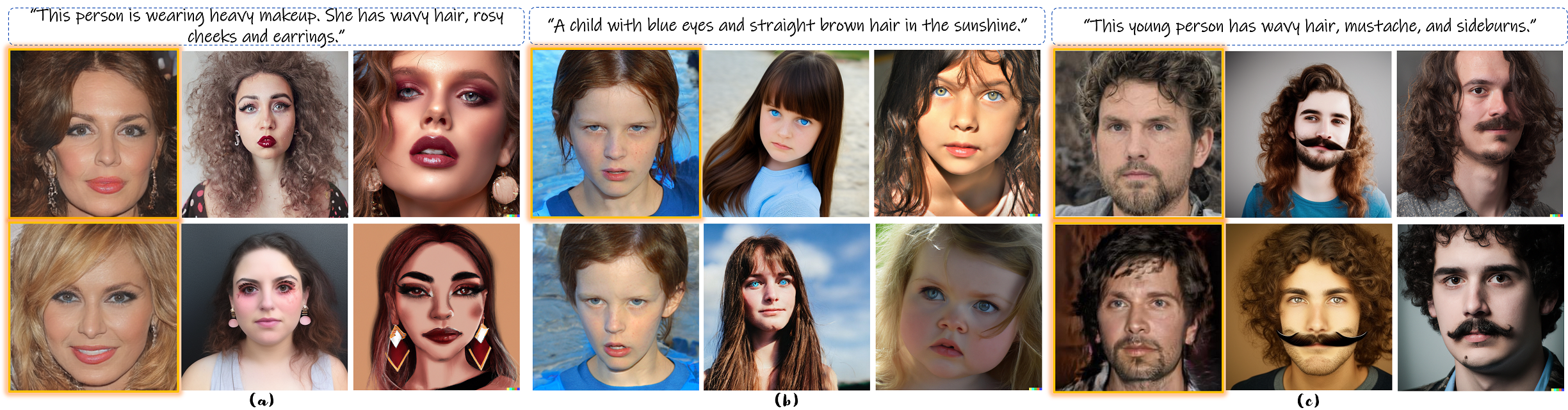}
  \caption{\textbf{Diverse text-to-face results generated from GAN-based~\cite{zhou2021generative,pinkney2022clip2latent,li2022stylet2i} / Diffusion-based~\cite{rombach2022high} / Transformer-based~\cite{ramesh2022hierarchical} models.} Images in orange boxes are captured from original papers (a)~\cite{zhou2021generative}, (b)~\cite{pinkney2022clip2latent} and (c)~\cite{li2022stylet2i}; others are generated by a pre-trained model~\cite{pinkney2022clip2latent} [(b) left bottom row], Dreamstudio [(a-c) middle row] and DALL-E 2 [(a-c) right row] online platforms from textual descriptions. Further details about online platforms are mentioned in \cref{BA}.} 
  \label{fig:T2F}
  \vspace{-2mm}
\end{figure*}

\textbf{{Text-to-Face}}\label{T2F} 
Text-to-image technology has been extensively applied in various applications. However, compared to other text-to-image applications, such as text-to-object or text-to-scene synthesis, text-to-face synthesis has received comparatively less attention (see underlined in \cref{tab:T2I}). As an early text-to-image application, facial visual reconstruction involved artists sketching suspects to help eyewitnesses recall more details. This process depends heavily on accurate descriptions and requires artists to have extensive training in facial sketching techniques. 
Moreover, text-to-face synthesis has significant potential in human retrieval, public security, and missing child renderings. Consequently, developing text-to-face synthesis techniques requires a meticulous approach to address challenges stemming from ambiguous and complex textual descriptions, thereby enhancing its practical significance. Comparison results for several text-to-face synthesis models are illustrated in \cref{fig:T2F}.

%-------------------------------------------------------------------------
\subsection{Text-to-X} \label{sec:t2x}

\begin{figure*}[t]
  \centering
  \includegraphics[width=0.96\textwidth,clip]{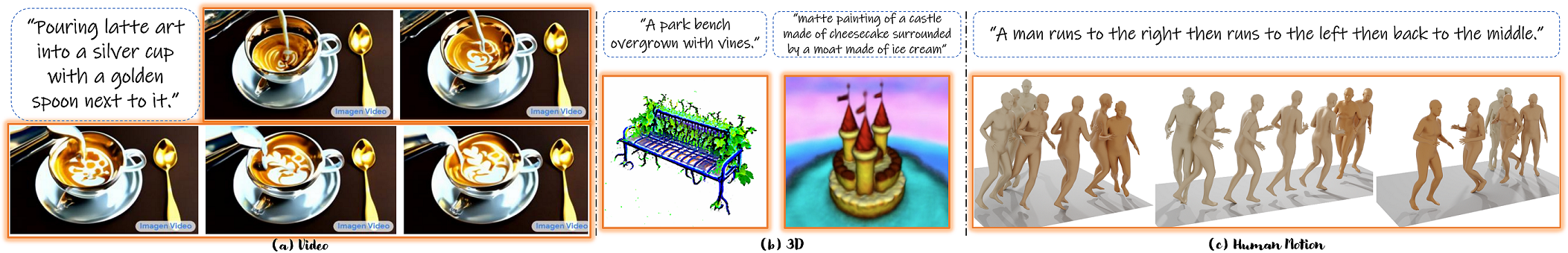}
  \caption{\textbf{Selected representative samples on Text-to-X.} Images are captured from original papers ((a)~\cite{ho2022imagen}, (b)-Left~\cite{xu2022dream3d}, (b)-Right~\cite{poole2022dreamfusion}, (c)~\cite{tevet2022human}) and remade.} 
  \label{fig:t2x}
\end{figure*}

\textbf{{Video}}\label{video} 
Automatically generating videos based on textual descriptions is a highly challenging task, as it requires addressing two critical issues: visual quality and semantic consistency. Visual quality refers to the generation of videos that are both realistic and visually coherent, while semantic consistency refers to the generated video sequence that can accurately represent the textual description.
%Text-to-video generation has numerous potential applications, including animation, film production, virtual reality, \etc. Researchers have developed various approaches to tackle these challenges. These models aim to capture not only the spatial information within individual frames but also the temporal dynamics that ensure the generated video sequences adhere to the given textual description while maintaining visual continuity and coherence across frames. 

Early text-to-video generation works based on GANs~\cite{pan2017create,li2018video,li2019storygan,deng2019irc,balaji2019conditional,kim2020tivgan} are limited to generating low-resolution short videos depicting simple scenes. With the ongoing development of transformer and diffusion models, recent methods based on transformers~\cite{hong2022cogvideo,villegas2022phenaki} and diffusion~\cite{singer2022make,ho2022imagen} improve the video quality (see \cref{fig:t2x} (a)) and speed~\cite{zhou2022magicvideo}. Uriel \etal~\cite{singer2023text} first propose a method for generating three-dimensional (3D) video from text descriptions. These dynamic scenes can be viewed from any camera location and composited into any 3D environment, breaking barriers between text-to-video and text-to-3D.

\textbf{{3D}}\label{3d}
Recent studies~\cite{lin2022magic3d,poole2022dreamfusion,wang2022score} have shown that neural radiance fields (NeRFs)~\cite{mildenhall2021nerf} can be effectively optimized using text-to-image diffusion models, leading to the generation of high-quality 3D content from textual descriptions (see \cref{fig:t2x} (b)-Left and \cref{fig:t2x} (b)-Right).
Additionally, Amit \etal~\cite{raj2023dreambooth3d} combine recent developments in personalizing text-to-image models (DreamBooth~\cite{ruiz2022dreambooth}) with text-to-3D generation (DreamFusion~\cite{poole2022dreamfusion}) and present DreamBooth3D, a text-to-3D generative model that can produce high-quality 3D images, which trained from as few as 3-6 casually captured images.

\textbf{{Human Motion}}\label{motion}
In the text-to-X domain, text-to-human motion generation~\cite{guo2022generating,tevet2022human,zhang2022motiondiffuse,petrovich2022temos} remains a relatively niche field. The generated motions are expected to have diversity, enabling exploration of the text-grounded motion space. More importantly, these motions should be accurately captured and well-adapted to the content from the input textual descriptions (see \cref{fig:t2x} (c)).

%-------------------------------------------------------------------------

\begin{figure*}[t]
  \centering
  \includegraphics[width=0.96\textwidth,clip]{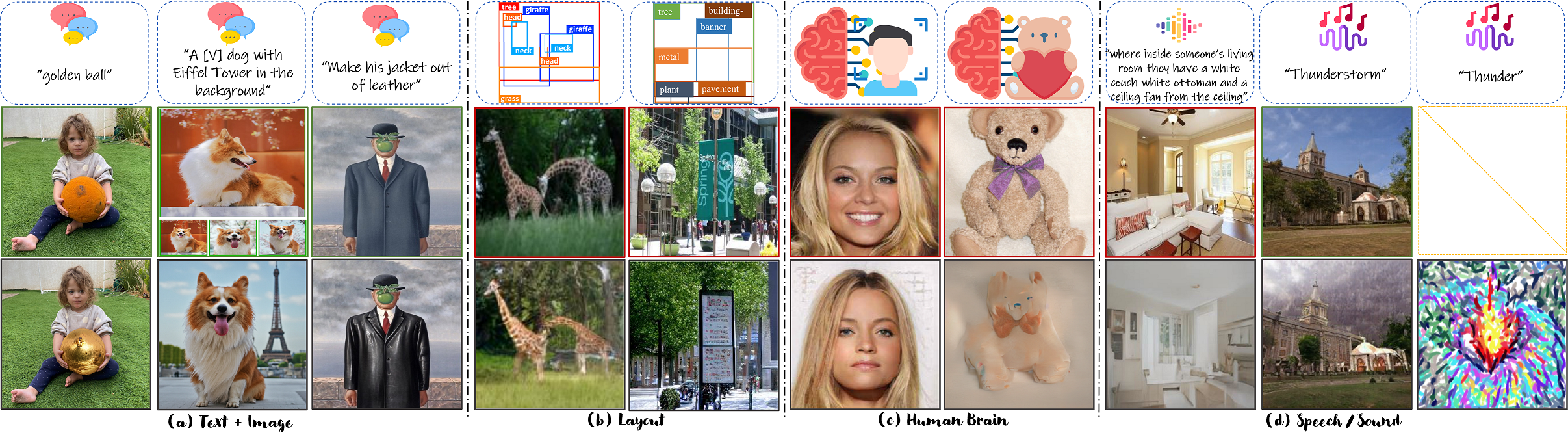}
  \caption{\textbf{Selected representative samples on X-to-Image.} Images are captured from original papers and remade. \textbf{(a)} Layered Editing~\cite{bar2022text2live} (Left), Recontextualization~\cite{ruiz2023dreambooth} (Middle), Image Editing~\cite{brooks2022instructpix2pix} (Right). \textbf{(b)} Context-Aware Generation~\cite{he2021context} (Left),  Model Complex Scenes~\cite{yang2022modeling} (Right). \textbf{(c)} Face Reconstruction~\cite{dado2022hyperrealistic} (Left), High-resolution Image Reconstruction~\cite{takagi2022high} (Right). \textbf{(d)} Speech to Image~\cite{wang2021generating} (Left), Sound Guided Image Manipulation~\cite{lee2022robust} (Middle), Robotic Painting~\cite{misra2023robot} (Right). \textbf{\textit{Legend:}} \textcolor{SORA}{\textit{X excluding “Additional Input Image”}} (Blue dotted line box, top row). \textcolor{NAE}{\textit{Additional Input Image}} (Green box, middle row). \textcolor{red}{\textit{Ground Truth}} (Red box, middle row). \textcolor{black}{\textit{Generated / Edited / Reconstructed Image}} (Black box, bottom row). } 
  \label{fig:x2i}
\end{figure*}

\subsection{X-to-Image} 
\label{sec:x2i}

\textbf{{Text + Image}}\label{t_plus_i}
Text-to-image generation and text-guided image generation (text+image-to-image) are two distinct techniques within image synthesis. Text-to-image generation models directly create images from a textual description. In this process, the synthesis model is trained to produce images that closely align with the input description, and the generated images are expected to represent the content of the input text visually.

In contrast, text-guided image generation refers to a process where a pre-existing image is modified or manipulated based on a textual input, utilizing textual information to make targeted modifications rather than generating a new image. This approach often involves tasks such as image editing~\cite{bar2022text2live,brooks2022instructpix2pix,huang2023region,pernuvs2023fice,couairon2022diffedit,zhu2022one,kim2022diffusionclip,avrahami2022blended,patashnik2021styleclip,nam2018text} (Samples are shown in \cref{fig:x2i} (a)-Left \& Right), recontextualization~\cite{ruiz2023dreambooth} (Results are shown in \cref{fig:x2i} (a)-middle), inpainting~\cite{ni2022n}, colorization~\cite{ghosh2022tic}, video generation~\cite{esser2023structure,fu2022tell,hu2022make,esser2023structure}, image stylization~\cite{esser2023structure,fu2022tell} or style transfer~\cite{fu2022language,jin2022language,kwon2022clipstyler,schaldenbrand2022styleclipdraw}, wherein the textual guidance serves to augment or refine specific elements of the initial image.
These advances in text-guided image generation have the potential to revolutionize the way we create and manipulate images, offering new possibilities for digital art, advertising and entertainment.

\textbf{{Layout}}\label{layout} 
Layout-to-image generation is a subdomain within image synthesis which focuses on generating coherent and visually consistent images based on a given layout. The layout typically consists of a structured representation of the spatial arrangement of various objects, scenes, or semantic regions within the image~\cite{he2021context,yang2022modeling,wang2022interactive,jahn2021high}. It includes information about the location, size, shape, and relationships of these elements. ~\cite{he2021context} introduces a context-aware feature transformation module in the generator to ensure the generated encoding accounts for coexisting objects in the scene (see \cref{fig:x2i} (b)-Left).
~\cite{yang2022modeling} compresses RGB images into patch tokens and only focuses on highly-related patch tokens specified by the spatial layout for modeling, thereby achieving disambiguation during the training process (see \cref{fig:x2i} (b)-Right).

\begin{table*}[]
\centering
\fontsize{7}{9.6}\selectfont
\caption{\textbf{Overview of various existing works with multi-modal tasks.}}
\begin{tabular}{cl||cc|ccccc|cccc}
\toprule[1.5pt]
\small \textbf{Year} & \diagbox {\small \textbf{Method}}{\small \textbf{Tasks}} &  T2I & T2V & (T+X)2I & LYT2I & SKT2I & SEG2I & I2I & UIG & SR & IC & Other Tasks\\
\midrule
\cellcolor{HIWA!45} 2021 & UFC-BERT~\cite{zhang2021ufc} & \usym{2714} & - & Partial Image & - & - & - & \usym{2714} & \usym{2714} & - & - & Multimodal Controls\\
\cellcolor{HIWA!45} 2021 & ERNIE-ViLG~\cite{zhang2021ernie} & \usym{2714} & - & - & - & - & - & - & - & - & \usym{2714} & Generative VQA\\

\cellcolor{KOKE!70} 2022 & OFA~\cite{wang2022ofa} & \usym{2714} & - & - & - & - & - & - & - & - & \usym{2714} & VQA, ...\\
\cellcolor{KOKE!70} 2022 & Frido~\cite{fan2022frido} & \usym{2714} & - & - & \usym{2714} & - & - & - & \usym{2714} & - & - & SG2I\\
\cellcolor{KOKE!70} 2022 & LDMs~\cite{rombach2022high} & \usym{2714} & - & - & \usym{2714} & - & - & - & - & \usym{2714} & - & Inpainting\\
\cellcolor{KOKE!70} 2022 & NÜWA~\cite{wu2022nuwa} &  \usym{2714} & \usym{2714} & Image/Video & - &  \usym{2714} & - & \usym{2714} & - & - & - & Video Prediction, ...\\
\cellcolor{KOKE!70} 2022 & MMVID~\cite{han2022show} & - & \usym{2714} & Partial Image & - & - & - & - & - & - & - & Multimodal Controls\\
\cellcolor{KOKE!70} 2022 & PoE-GAN~\cite{huang2022multimodal} & \usym{2714} & - & SEG/SKT/Image & - & \usym{2714} & \usym{2714} & - & - & - & - & (SEG+SKT)2I\\
\cellcolor{KOKE!70} 2022 & AugVAE-SL~\cite{kim2022verse} & \usym{2714} & - & - & - & - & - & - & - & - & \usym{2714} & Image Reconstruction\\
\cellcolor{KOKE!70} 2022 & NUWA-Infinity~\cite{wu2022nuwain} & \usym{2714} & \usym{2714} & - & - & - & - & - & \usym{2714} & - & - & Outpainting(HD), ...\\
%VD~\cite{xu2022versatile} & \usym{2714} & - & Image & - & - & - & \usym{2714} & - & - & \usym{2714} & Editable I2T2I & 2022\\

\cellcolor{KOKE!85} 2023 & SDG~\cite{liu2023more} & \usym{2714} & - & Image & - & - & - & \usym{2714} & - & - &  - & Style-guided, ...\\
\cellcolor{KOKE!85} 2023 & Muse~\cite{chang2023muse} & \usym{2714} & - & Image & - & - & - & - & - & - & - & Inpainting, Outpainting\\
\cellcolor{KOKE!85} 2023 & MCM~\cite{ham2023modulating} & - & - & SEG/SKT & - & \usym{2714} & \usym{2714} & - & - & - & - & (SEG+SKT)2I\\
\cellcolor{KOKE!85} 2023 & TextIR~\cite{liu2023more} & - & - & Image & - & - & - & - & - & \usym{2714} & - & Inpainting, Colorization\\
\cellcolor{KOKE!85} 2023 & GigaGAN~\cite{Kang2023ScalingUG} & \usym{2714} & - & Image & - & - & - & - & - & \usym{2714} & - & Style Mixing, ...\\
\cellcolor{KOKE!85} 2023 & UniDiffuser~\cite{bao2022one} & \usym{2714} & - & Image & - & - & - & - & \usym{2714} & - & \usym{2714} & Joint Generation\\
\cellcolor{KOKE!85} 2023 & Visual ChatGPT~\cite{wu2023visual} & \usym{2714} & - & Image & - & \usym{2714} & \usym{2714} & \usym{2714} & - & - & \usym{2714} & Edge-to-Image,...\\
\bottomrule[1.5pt]
\end{tabular}
\vspace{0.5mm}
\raggedright
\footnotesize{{\underline{<\textbf{Acronym}: Meaning>}} \textbf{T2I}:Text-to-Image; \textbf{T2V}:Text-to-Video; \textbf{T+X}:Text+X; \textbf{LYT}:Layout;  \textbf{SKT}:Sketch; \textbf{SEG}:Segmentation; \textbf{UIG}:Unconditional Image Generation;  \textbf{SR}:Super Resolution; \textbf{IC}:Image Captioning/Image-to-Text; \textbf{VQA}:Visual Question Answering; \textbf{HD}:High Definition; \textbf{SG}:Scene Gragh.}
\label{tab:multi_tasks}
\vspace{-2mm}
\end{table*}

\textbf{{Human Brain}}\label{brain}
Neural encoding and decoding are two fundamental concepts in neuroscience to make sense of brain-activity data. The former investigates how individual neurons or neural networks represent information through action potentials. The latter involves mapping brain responses to sensory stimuli through feature space. Both fields have a long history of research~\cite{kriegeskorte2006information,jain2023selectivity} and are still actively studied with the development of deep learning~\cite{bermudez2018learning,shen2019deep,shen2019end}. 

Dado \etal~\cite{dado2022hyperrealistic} integrate GANs~\cite{goodfellow2014generative} in the neural decoding of faces. They utilize a pretrained generator of progressive growing GAN (PGGAN)~\cite{karras2017progressive} to synthesize photorealistic faces from latents. Meanwhile, a decoding model predicts latents from whole-brain functional Magnetic Resonance Imaging (fMRI) activations (as shown in \cref{fig:x2i} (c)-Left). 
% that were then fed to the generator for reconstruction or synthesis. The efficacy of this approach was confirmed through the production of valid reconstructions of visual perception 
%However, these reconstructions contain certain biases, such as young, western-looking faces without eyeglasses tend to be overrepresented due to the image statistics of the celebrity training set. 
Takagi and Nishimoto~\cite{takagi2022high} demonstrate the ability to decode perceptual content from brain activity by converting it into internal representations of Stable Diffusion~\cite{rombach2022high} without fine-tuning (as shown in \cref{fig:x2i} (c)-Right). %Furthermore, they evaluate the correspondence between brain function and the individual components of the model by converting the internal representations of the Stable Diffusion model back into brain activity via encoding.

\textbf{{Speech / Sound}}\label{speech}
According to the statistics, approximately 50\% of the world's languages lack a written form, which makes it impossible to benefit from current text-based technologies. Speech-to-image generation approaches~\cite{wang2021generating,lee2022robust,misra2023robot,li2020direct} translate speech descriptions into photorealistic images without relying on textual information. The speech embedding network in ~\cite{wang2021generating} learns speech embedding with visual supervision, while the generative model synthesizes visually consistent images (see \cref{fig:x2i} (d)-Left) based on corresponding speech descriptions.

In addition to the image generation from speech, certain studies also investigate the provision of targeted audio between distinct modalities, such as sound-image~\cite{lee2022robust,misra2023robot} (see \cref{fig:x2i} (d)-Middle \& Right), speech-gesture~\cite{kucherenko2021speech2properties2gestures,ao2022rhythmic}, music-motion~\cite{aristidou2021rhythm}, music-dance~\cite{chen2021choreomaster,siyao2022bailando}, \etc.

%[Optional]-------------------------------------------------------------------------
% \subsection{Downstream Tasks} 
% \label{sec:downstream}
% \textbf{\textit{Visual Retrieval:}}
% \label{VR}
% \textbf{\textit{Style Transfer:}}
% \label{ST}
%-------------------------------------------------------------------------

\subsection{Multi Tasks} 
\label{sec:multi}

Multimodal learning, incorporating multiple data sources, including text, image, audio, and video,  has gained increasing attention due to its ability to solve multiple related tasks simultaneously. Additionally, multi-task learning enables the models to share knowledge across tasks, resulting in improved performance on individual tasks and enhanced generalization capabilities. 

In recent years, significant progress has been made in developing multimodal and multi-task learning approaches, as shown in \cref{tab:multi_tasks}. These architectures facilitate the integration of various modalities by learning shared representations and task-specific features, ultimately leading to more effective and efficient models.

%Multimodal learning with multi-tasks has broad applicability across numerous domains, such as computer vision, natural language processing, and speech processing. Researchers can develop more comprehensive models capable of tackling complex and interrelated tasks, pushing the boundaries of what machine learning systems can achieve.

%#########################################################################
\section{Discussion}
\label{sec:discussion}
%-------------------------------------------------------------------------
\subsection{Compounding Issues}
\label{sec:issue}

\textbf{{Computational Aesthetic}}\label{AQ}
As AI technology continues to evolve and become more accessible, users are increasingly seeking convenience, better user experience, visual aesthetics, and perceived attractiveness. Computational aesthetic evaluation has been employed for assessing potential AI creativity is much older than text-to-image generation~\cite{galanter2012computational}. State-of-the-art text-to-image models increasingly consider aesthetics which attempt to produce better images via better prompts~\cite{oppenlaender2022taxonomy}.

Since computational aesthetic evaluation remains challenging, human feedback is necessary to determine the optimal prompt formulation and keyword combinations. Pavlichenko and Ustalov~\cite{pavlichenko2022best} adopt the most popular keywords to evaluate their effectiveness in Stable Diffusion~\cite{rombach2022high} and propose a set of keywords that produce images with the highest degree of aesthetic appeal. Jonas~\cite{oppenlaender2022creativity} emphasizes that the ability to generate effective prompts depends on a thorough understanding of the training set and knowledge of various prompt modifiers. This ability and knowledge together form the field of “prompt engineering”. 

\textbf{{Prompt Engineering}}\label{PE}
In NLP, prompt-based learning (PBL)~\cite{zhou2022learning,zhou2022conditional,liu2023pre} employ pre-trained language models on large amounts of text data to address diverse downstream tasks. The process of prompt engineering~\cite{reynolds2021prompt} is crucial for generating task-specific prompt templates, and the effectiveness of PBL greatly depends on the construction method of these templates. The most basic prompt construction method is manual construction, which involves designing appropriate text templates for the target problem, such as translation~\cite{brown2020language}. %The construction method of prompt templates significantly impacts the effect, which is a crucial factor for the success of PBL. conditional text generation~\cite{schick2020few}

\begin{figure}[t]
  \centering
  \includegraphics[width=0.49\textwidth,clip]{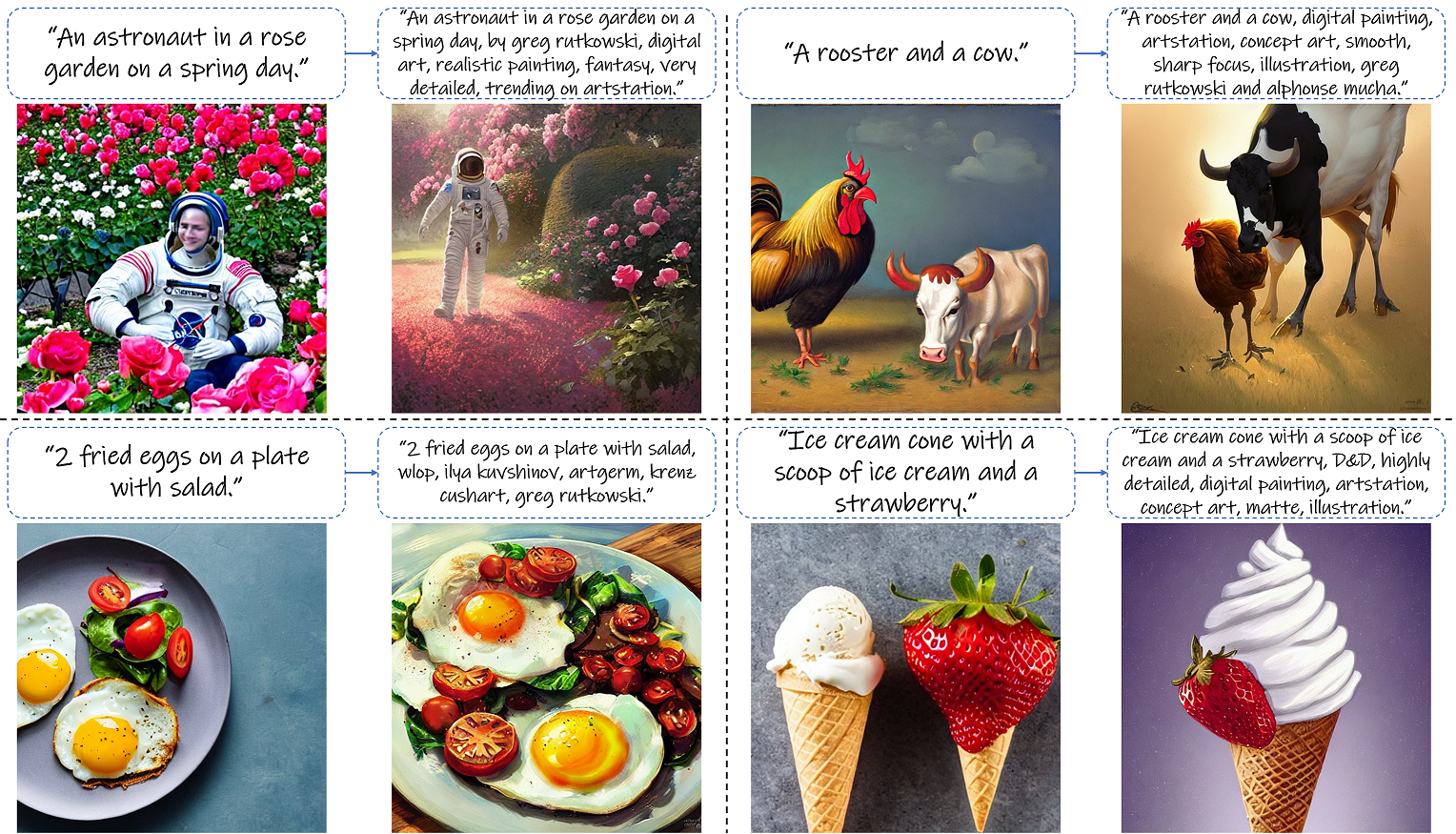}
  \caption{\textbf{Sample images generated from user input and optimized prompts.} Images are generated by Stable Diffusion~\cite{rombach2022high}, while optimized prompts are generated by Promptist~\cite{hao2022optimizing}.} 
  \label{fig:PE}
  \vspace{-2mm}
\end{figure}

In CV, text-to-image generative models present a powerful way to generate realistic images. While using text as input allows for an unlimited range of outputs, users must engage in trial and error with the text prompt when the output is poor quality. Prompt engineering~\cite{liu2022design} for text-to-image, also referred to as prompt design~\cite{reynolds2021prompt} or prompting, is an emerging technology that utilizes carefully selected sentences to achieve a specific visual style in the synthesized image~\cite{rombach2022text}. Liu and Chilton~\cite{liu2022design} design five experiments to explore different aspects of prompt engineering for text-to-image generative models, including prompt permutations, random seeds, optimization length, style keywords, and subject and style keywords. They also perform a thorough analysis of the failure and success modes of the above generations. Hao \etal~\cite{hao2022optimizing} propose a prompt adaptation framework (called “Promptist”) as an alternative to laborious manual prompt engineering. They use a few manual prompts for supervised fine-tuning and reinforcement learning to generate better prompts. Reinforcement learning involves a reward function that encourages generating more aesthetically appealing images while maintaining the original user intentions. Experimental results (as shown in \cref{fig:PE}) indicate that optimized prompts can enhance performance in the following aspects: “Aesthetics augmentation” (top-left), “Content rationalization” (top-right), “Style transformation” (bottom-left), and “Accurate expression” (bottom-right).

%-------------------------------------------------------------------------
\subsection{Business Analysis}
\label{sec:business}

\textbf{{Online Platforms}}\label{BA}
As AI-based image generators become a topic of widespread discussion, significant technological advancements have greatly enhanced the accessibility of these tools to the general populace. The emergence of OpenAI's DALL-E~\cite{ramesh2021zero} and DALL-E 2~\cite{ramesh2022hierarchical} marked a crucial milestone in the evolution of AI-based text-to-image generation.
While certain text-to-image tools are available for free, others may require a subscription or offer a trial period. Moreover, several platforms provide additional features to enhance the user experience. \Cref{tab:BA} provides an overview of represented online platforms and websites that easily create images from text prompts.

\begin{table}[]
\centering
\fontsize{5.2}{8.5}\selectfont
\caption{\textbf{Overview of represented online platforms for text-to-image generation.} (Arrange in ascending order of usability and complexity.) DALL-E 2~\cite{ramesh2022hierarchical} and Stable Diffusion~\cite{rombach2022high} are the two most commonly utilized models.}
\begin{tabular}{c||cccc}
\toprule[1.5pt]
\textbf{Platform} & \textbf{Models} & \textbf{Price} & \textbf{Additional} & \textbf{Links} \\
\midrule
\cellcolor{JINZAMOMI!30} DeepAI & ~\cite{rombach2022high} & Free & Style & \href{https://deepai.org/machine-learning-model/text2img}{Website}\\
\cellcolor{JINZAMOMI!30} Wombo & - & Free & Style & \href{https://app.wombo.art/}{Website}\\
\cellcolor{JINZAMOMI!30} Craiyon & - & Free & - & \href{https://www.craiyon.com/}{Website}\\
\cellcolor{JINZAMOMI!30} Yunjing & ~\cite{rombach2022high}, ... & Free & Chinese prompts & \href{https://yunjing.gallery/}{Website}\\
\cellcolor{JINZAMOMI!30} Replicate & ~\cite{rombach2022high} & Free & - & \href{https://replicate.com/stability-ai/stable-diffusion}{Website}\\
\cellcolor{JINZAMOMI!30} Nightcafe & ~\cite{ramesh2022hierarchical},~\cite{rombach2022high}, ... & Free & Style & \href{https://creator.nightcafe.studio/}{Website}\\
\cellcolor{JINZAMOMI!30} Freehand & - & Free & Chinese prompts & \href{https://freehands.cn/}{Website}\\
\cellcolor{JINZAMOMI!30} HuggingFace & ~\cite{rombach2022high} & Free & - & \href{https://huggingface.co/spaces/stabilityai/stable-diffusion}{Website}\\
\cellcolor{JINZAMOMI!30} SD Playground & ~\cite{rombach2022high} & Free & - & \href{https://stablediffusionweb.com/}{Website}\\
\cellcolor{JINZAMOMI!30} Bing Image Creator & ~\cite{ramesh2022hierarchical} & Free & - &  \href{https://www.bing.com/create}{Website}\\
\cellcolor{JINZAMOMI!40} Lexica & - & Monthly 100 images & Search & \href{https://lexica.art/}{Website}\\
\cellcolor{JINZAMOMI!55} starryai & - & Daily 5 free credits & Style, Image+Text & \href{https://create.starryai.com/my-creations}{Website}\\
\cellcolor{JINZAMOMI!65} Dreamstudio & ~\cite{rombach2022high} & 200 free credits & Image+Text & \href{https://beta.dreamstudio.ai/}{Website}\\
\cellcolor{JINZAMOMI!80} Midjourney & - & 20 times free & - & \href{https://www.midjourney.com/}{Website}\\
\cellcolor{JINZAMOMI!95} DALL-E 2 & ~\cite{ramesh2022hierarchical} & Monthly 15 free credits & - & \href{https://openai.com/product/dall-e-2}{Website}\\
\cellcolor{JINZAMOMI!95} Firefly  & - & Application is required & - & \href{https://firefly.adobe.com/generate/images}{Website}\\
\bottomrule[1.5pt]
\end{tabular}
\label{tab:BA}
\vspace{-2mm}
\end{table}

\textbf{{Ethical Considerations}}\label{EC}
Despite the significant progress made in open-source text-to-image generation models, the technology has not yet reached commercial viability due to concerns about unconsciously producing offensive or potentially dangerous biased images. These biases include ambiguity, immorality, stereotypes, or other negative connotations. This is frequently attributed to a lack of consideration for ethical considerations in conventional approaches.

For \textbf{\textit{ambiguity issues}}, Mehrabi \etal~\cite{mehrabi2022elephant} present a Text-to-image Ambiguity Benchmark (TAB) dataset and a disambiguation framework for generating images that more closely align with user intention, as well as novel automatic evaluation procedures for assessing ambiguity resolution. They employ few-shot learning with specific language models to disambiguate ambiguous prompts with the aid of human feedback.
For \textbf{\textit{immoral issues}}, Park \etal~\cite{park2022judge} first propose effective ethical image manipulation methods by localizing immoral attributes, including blurring, inpainting, and text-driven image manipulation, that have demonstrated effectiveness.
For \textbf{\textit{stereotypes issues}}, Struppek \etal~\cite{struppek2022biased} demonstrate that text-based image generation models are sensitive to character encodings, with the insertion of even a single homoglyph at an arbitrary position can introduce cultural biases and stereotypes into the generated images, thereby influencing the image generation process. Additionally, Bansal \etal~\cite{bansal2022well} indicate that certain keywords, such as `irrespective of gender' and `culture', can trigger substantial variations and diversities in model predictions, particularly in the context of gender bias within ethical interventions. Federico \etal~\cite{bianchi2022easily} investigate accessible text-to-image generation models and expose the extent of categorization, stereotyping, and complex biases in the Stable Diffusion model~\cite{rombach2022high} and generated images.

%-------------------------------------------------------------------------
\subsection{Challenges \& Future Outlooks}
\label{sec:future}

\begin{minipage}{0.02\textwidth}
\includegraphics[width=\linewidth]{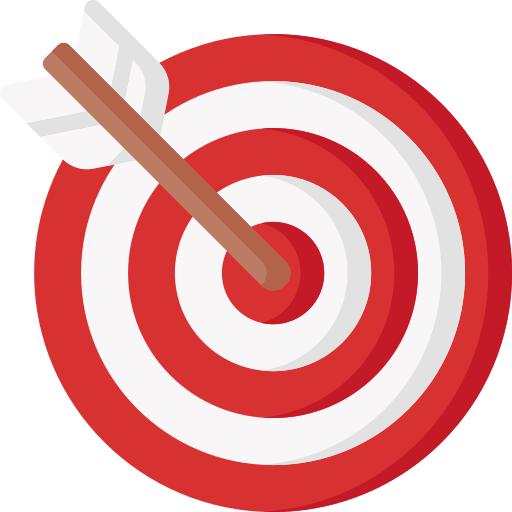}
\end{minipage}
As mentioned in \cref{sec:dataset}, empirical evidence has demonstrated that supervised learning models are highly effective in accomplishing tasks for which they have been specifically trained by leveraging labeled data. However, their performance tends to decline when confronted with tasks beyond their range, as their proficiency is heavily dependent on the quality of the labeled data. Furthermore, it is impractical to label every piece of information available worldwide.
Consequently, there has been a growing emphasis on developing more versatile, generalist models within the field of AI. The aim is to develop models capable of performing well across a range of tasks, thereby reducing dependence on vast quantities of labeled data.

It is noteworthy that the existing text-to-face datasets suffer from a lack of large-scale image-text pairs, which inadequacy significantly impedes progress in automatic text-to-face synthesis research. The primary reason lies in the burdensome process of collecting and annotating facial images. Furthermore, unlike other image categories, such as birds or flowers, facial features are more complex and multifaceted, including numerous factors such as ethnicity, gender, age, expression, and environmental context~\cite{zhou2021}. Therefore, developing large-scale text-to-face datasets entails greater challenges than those associated with other image domains.

\begin{minipage}{0.02\textwidth}
\includegraphics[width=\linewidth]{pic/target.png}
\end{minipage}
As discussed in \cref{sec:t2i}, text-to-face technology can potentially obtain valuable support from eyewitness testimony. However, the reliability of such testimony may be compromised due to factors such as fear or cognitive limitations,  leading to inconsistencies between the provided description and the suspect's actual physical appearance. Therefore, the ability to manipulate specific visual features in synthetic facial images while maintaining other attributes consistent with the input description has become increasingly crucial in developing text-to-face synthesis technology for the public safety domain. 

This necessitates an in-depth investigation aimed at improving the effectiveness of text-to-face synthesis approaches while adhering to the following specific compliance requirements:
(1) \textit{Discriminability}: Ensuring that the generated images are recognizable as individual persons.
(2) \textit{High Resolution}: Producing images with high resolution to facilitate detailed analysis.
(3) \textit{Photorealism}: Generating images that closely resemble authentic faces.
(4) \textit{Diversity}: Creating a variety of images from multiple viewpoints.
(5) \textit{Fidelity}: Ensuring that the generated images are consistent with the input description.
(6) \textit{Controllability}: Enabling selective manipulation of different attributes with text prompts while preserving other irrelevant attributes.

\begin{minipage}{0.02\textwidth}
\includegraphics[width=\linewidth]{pic/target.png}
\end{minipage}
As mentioned in \cref{sec:t2x} and \cref{sec:x2i}, text-to-X and X-to-image tasks are subcategories of multimodal learning. Here are some challenges:
(1) \textit{Alignment}: Ensuring that different modalities are appropriately aligned so that the generated results accurately reflect the input modality.
(2) \textit{Data scarcity}: Collecting and annotating large-scale multimodal datasets is time-consuming and expensive, particularly for specialized domains, which limit the performance of existing models.
(3) \textit{Scalability}: Developing models that efficiently handle large-scale multimodal data without compromising performance is an ongoing challenge. This includes addressing the memory and computational requirements associated with managing multiple modalities.

\begin{minipage}{0.02\textwidth}
\includegraphics[width=\linewidth]{pic/target.png}
\end{minipage}
Despite the use of thoughtfully designed prompts to promote diversification and subvert undesired stereotypes, the limitations of text-to-image generation models are evident. As discussed in \cref{sec:issue} and \cref{sec:business}, several instances have illustrated the fragility of these models, despite their remarkable ability to generate images. This issue remains unsolved and requires significant further research.

Universal access and commercial use of open-source AI generators have been a noticeable trend, which may have both positive and negative consequences. Some may contend that regardless of its quality, AI technology should be accessible to everyone. Nevertheless, the long-term implications of this trend remain uncertain and require further evaluation and consideration.

%#########################################################################
\section{Conclusion}
\label{sec:conclusion}
Situated at the intersection of theoretical foundations and practical applications, multimodal learning has reached a critical juncture in today's rapidly evolving landscape. In recognition of the importance of synthesizing multiple domains, our objective in this review is to provide a comprehensive and scholarly analysis of contemporary advancements in multimodal learning. This survey aims to seamlessly connect theoretical underpinnings with real-world considerations by analyzing and summarizing various vision and language methodologies, innovative techniques, and emerging trends. Ultimately, this work seeks to build an integrated framework and highlight current challenges and opportunities that facilitate a deeper understanding of multimodal learning in the research community and its real-world implications.

%%%%%%%%% REFERENCES
{\small
\bibliographystyle{ieee_fullname}
\bibliography{Mybib}
}

\end{document}